%% file: main.tex
\documentclass[10pt,conference,a4paper]{IEEEtran}
\ifCLASSOPTIONcompsoc
  \usepackage[nocompress]{cite}
\else
  \usepackage{cite}
\fi

\ifCLASSINFOpdf
\else
\fi
\usepackage{todonotes}

\usepackage{multirow}
\usepackage[inline]{enumitem}
\usepackage{arydshln}
\usepackage{booktabs}
\usepackage{colortbl}

\usepackage{subcaption}

\makeatletter
\def\blfootnote{\xdef\@thefnmark{}\@footnotetext}
\makeatother

\input{utils}

\begin{document}
% \bstctlcite{IEEEexample:BSTcontrol}

\title{Automated Pruning for Deep Neural Network Compression}

% \author{
%   \IEEEauthorblockN{Franco Manessi}
% \IEEEauthorblockA{Strategic Analytics\\ lastminute.com group\\ Chiasso, Switzerland \\Email: franco.manessi@lastminute.com}
% \and
% \IEEEauthorblockN{Alessandro Rozza}
% \IEEEauthorblockA{Strategic Analytics\\ lastminute.com group\\ Chiasso, Switzerland \\
% Email: alessandro.rozza@lastminute.com}
% \and
% \IEEEauthorblockN{Simone Bianco}
% \IEEEauthorblockA{DISCo \\
% Universit\`a degli Studi di Milano Bicocca\\
% Milano, Italy\\
% Email: simone.bianco@disco.unimib.it}
% \and
% \IEEEauthorblockN{Paolo Napoletano}
% \IEEEauthorblockA{DISCo \\
% Universit\`a degli Studi di Milano Bicocca\\
% Milano, Italy\\
% Email: paolo.napoletano@disco.unimib.it}
% \and
% \IEEEauthorblockN{Raimondo Schettini}
% \IEEEauthorblockA{DISCo \\
% Universit\`a degli Studi di Milano Bicocca\\
% Milano, Italy\\
% Email: raimondo.schettini@disco.unimib.it}
% }

% \author{\IEEEauthorblockN{Franco Manessi\IEEEauthorrefmark{1}, 
% Alessandro Rozza\IEEEauthorrefmark{1},
% Simone Bianco\IEEEauthorrefmark{2},
% Paolo Napoletano\IEEEauthorrefmark{2} and
% Raimondo Schettini\IEEEauthorrefmark{2}}
% \IEEEauthorblockA{\IEEEauthorrefmark{1}Strategic Analytics\\ lastminute.com group\\ Chiasso, Switzerland \\Email: \{first\_name.last\_name\}@lastminute.com\\ Equal contribution} \and 
% \IEEEauthorblockA{\IEEEauthorrefmark{2}DISCo \\
% Universit\`a degli Studi di Milano Bicocca\\
% Milano, Italy\\
% Email: \{first\_name.last\_name\}@disco.unimib.it}
% }

\author{\IEEEauthorblockN{Franco Manessi\IEEEauthorrefmark{1}, 
Alessandro Rozza\IEEEauthorrefmark{1},
Simone Bianco\IEEEauthorrefmark{2},
Paolo Napoletano\IEEEauthorrefmark{2} and
Raimondo Schettini\IEEEauthorrefmark{2}}
\IEEEauthorblockA{\IEEEauthorrefmark{1}Strategic Analytics --- lastminute.com group, Chiasso, Switzerland \\Email: \{first\_name.last\_name\}@lastminute.com\\ Equal contribution} 
\IEEEauthorblockA{\IEEEauthorrefmark{2}DISCo ---
Universit\`a degli Studi di Milano Bicocca,
Milano, Italy\\
Email: \{first\_name.last\_name\}@disco.unimib.it}
}


% compsoc conferences compared to traditional conferences), use this
% \author{\IEEEauthorblockN{Michael Shell\IEEEauthorrefmark{1},
% Homer Simpson\IEEEauthorrefmark{2},
% James Kirk\IEEEauthorrefmark{3}, 
% Montgomery Scott\IEEEauthorrefmark{3} and
% Eldon Tyrell\IEEEauthorrefmark{4}}
% \IEEEauthorblockA{\IEEEauthorrefmark{1}School of Electrical and Computer Engineering\\
% Georgia Institute of Technology,
% Atlanta, Georgia 30332--0250\\ Email: see http://www.michaelshell.org/contact.html}
% \IEEEauthorblockA{\IEEEauthorrefmark{2}Twentieth Century Fox, Springfield, USA\\
% Email: homer@thesimpsons.com}
% \IEEEauthorblockA{\IEEEauthorrefmark{3}Starfleet Academy, San Francisco, California 96678-2391\\
% Telephone: (800) 555--1212, Fax: (888) 555--1212}
% \IEEEauthorblockA{\IEEEauthorrefmark{4}Tyrell Inc., 123 Replicant Street, Los Angeles, California 90210--4321}}


\maketitle

\begin{abstract}
In this work we present a method to improve the pruning step of the current 
state-of-the-art methodology to compress neural networks. The novelty of the 
proposed pruning technique is in its differentiability, which allows pruning to 
be performed during the backpropagation phase of the network training. This 
enables an end-to-end learning and strongly reduces the training time. The 
technique is based on a family of differentiable pruning functions and a new 
regularizer specifically designed to enforce pruning. The experimental results 
show that the joint optimization of both the thresholds and the network weights 
permits to reach a higher compression rate, reducing the number of weights of 
the pruned network by a further 14\% to 33\% compared to the current 
state-of-the-art. Furthermore, we believe that this is the first study where the 
generalization capabilities in transfer learning tasks of the features extracted 
by a pruned network are analyzed. To achieve this goal, we show that the 
representations learned using the proposed pruning methodology maintain the same 
effectiveness and generality of those learned by the corresponding 
non-compressed network on a set of different recognition tasks.
\end{abstract}


% \fi
\blfootnote{Published as a conference paper at ICPR 2018.}

\IEEEpeerreviewmaketitle

 % \renewcommand{\thefootnote}{\fnsymbol{footnote}}
 % \footnotetext[2]{Equal contribution}
 % \renewcommand{\thefootnote}{\arabic{footnote}}

\input{introduction}
\input{related}
\input{method}
\input{results}
\input{cnnfeatures}
\input{conclusion}

\input{future}

\ifCLASSOPTIONcompsoc
  \section*{Acknowledgments}
\else
  \section*{Acknowledgment}
\fi
The Tesla K40 used for this research was donated to University of 
Milano-Bicocca by the NVIDIA Corporation.



\bibliographystyle{IEEEtran}

\bibliography{IEEEfull,bibliography.bib}


% \end{thebibliography}

\end{document}

%% file: utils.tex
\usepackage{bm}
\usepackage{xspace}
\usepackage{mathrsfs}
\usepackage{mathtools}

\usepackage{amsmath}
\usepackage{amssymb}

\newcommand\Matrix[1]{\bm{#1}}
\newcommand\ie{i.e.\xspace}
\newcommand\eg{e.g.\xspace}
\newcommand\etal{et al.\xspace}
\newcommand\Reals{\mathbb{R}}
\DeclareMathOperator{\relu}{ReLU}
\newcommand\vect[1]{{\boldsymbol{#1}}}

\newcommand\Set[1]{{\bm{\mathcal{#1}}}}
\newcommand\muacro[1]{\texttt{#1}} 
\newcommand\Minus{\textrm{-}}
\newcommand\Norm[1]{\left\lVert#1\right\rVert}
\newcommand\Network[1]{#1}
\newcommand\Layer[1]{#1}

\newtheorem{definition}{Definition}

\newcommand\lenetff{\muacro{LeNet-300-100}\xspace}
\newcommand\lenetconv{\muacro{LeNet-5}\xspace}
\newcommand\alexnet{\muacro{AlexNet}\xspace}
\newcommand\mnist{{MNIST}\xspace}
\newcommand\ilsvrc{{ILSVRC-2012}\xspace}

%% file: introduction.tex
\section{Introduction}
In the last five years, deep neural networks have achieved state-of-the-art 
results in many computer vision tasks. A possible limitation of these approaches 
is related to the fact that these models are characterized by a large number of 
weights that consume considerable storage and memory resources.

The aforementioned drawback makes it difficult to deploy these models on embedded 
systems with limited hardware resources. Furthermore, running large neural 
networks requires a lot of memory bandwidth to fetch the weights and a lot of 
computation for matrix multiplication, which consume a considerable amount of 
energy. Moreover, considering the mobile market, the majority of the app-stores 
are particularly sensitive to the size of the binary files, potentially reducing 
the spread of big applications, which can be downloaded just using a WiFi 
connection (\eg if their size is greater that 100MB). 

To overcome these limitations, reducing the storage and energy requirements to 
run inference of these large networks also on mobile devices, many different 
approaches of network compression have been proposed. Among them, we mention: 
\begin{enumerate*}[label=(\roman*)]
    \item weight sharing;
    \item \emph{pruning} network connections whose corresponding weights are
        below some threshold;
    \item \emph{quantizing} network weights so to reduce the precision with 
        which they are stored;
    \item \emph{binarizing} networks by employing only two-valued weights.
\end{enumerate*}

Han~\etal presented in \cite{Han2015} an interesting approach called \emph{Deep 
Compression}, which is able to reduce the storage requirements of neural 
networks without affecting their accuracy. This framework:  
\begin{enumerate*}[label=(\roman*)]
    \item prunes the network by learning only the important connections; 
    \item it quantizes the weights to enforce weight sharing; 
    \item it applies Huffman coding.
\end{enumerate*}
 
The network pruning might be considered the most relevant part of this framework 
and is composed of the following steps: 
\begin{enumerate}[noitemsep,topsep=0pt, label=(\roman*)]
    \item it learns the connectivity via normal network training;
    \item it prunes the small-weight connections (i.e. all connections with 
        weights below a threshold);\label{step2}
    \item it retrains the network to learn the final weights for the remaining 
        sparse connections.\label{step3}
\end{enumerate}

The main limitation of this part is due to the fact that, to identify the 
appropriate threshold parameter value, this approach has to re-iterate steps 
\ref{step2} and \ref{step3} many times, wasting a lot of computational 
resources.  Moreover, since the threshold and the network weights are not 
jointly optimized during the training phase, this can produce a sub-optimal 
solution not able to achieve the maximum compression rate. 

We improved the pruning methodology of \emph{Deep Compression} by making it 
differentiable with respect to the threshold parameters. This allows to 
automatically estimate the best threshold parameters, together with the network 
weights, during the learning phase, thus strongly reducing the training time. 
This is due to fact that, we execute the learning phase only once instead of 
repeating the retraining of the network for each of the (many) tested threshold 
parameters (\ie step \ref{step3}).
This approach allows to overcome another limitation of \emph{Deep Compression}: 
Han~\etal's technique limits the exploding complexity of its iterative algorithm 
by seeking for \emph{one} threshold value shared by all the layers, which then 
are pruned according to threshold value and the standard deviation of their 
weights. This approach might lead to a sub-optimal pruned configuration compared
to a procedure finding a \emph{per-layer} threshold. The approach 
proposed in this paper is able to find \emph{layer-specific} thresholds thanks 
to its differentiable nature, thus avoiding the simplification needed by 
\emph{Deep Compression}. Moreover, our pruning technique is able to 
achieve better results considering the compression rate of the pruned model, 
obtaining a number of weights of the pruned networks that is $14\%$ to $33\%$ 
lower than the ones obtained by \emph{Deep Compression}.

It is important to underline that, in this paper, we focus only on the pruning 
stage of the \emph{Deep Compression} pipeline, since quantization is 
orthogonal to network pruning \cite{han2015learning} and it is known that 
pruning, quantization, and Huffman coding can compress the network 
without interfering each other \cite{Han2015}.

Since deep neural networks are very often used in transfer learning scenarios 
\cite{sharif2014cnn}, we investigate here if the representations learned using 
the proposed pruning methodology have the same effectiveness and generality of 
those learned by a non-compressed network. The transfer learning experiments 
are performed on different recognition tasks, such as  object image 
classification, scene recognition, fine grained recognition, attribute 
detection, and image retrieval. To the best of our knowledge, this is the first 
study where such experiments have been performed using the features extracted 
by a compressed network.

Summarizing, our main contributions are the following:
\begin{enumerate}[noitemsep,topsep=0pt,label=(\roman*)]
	\item an approach that automatically determines the threshold values of 
        the pruning phase in a differentiable fashion, reducing the training 
        time and achieving better compression results with no or negligible 
        drop in accuracy.
	\item the evaluation of the compressed network in terms of transfer learning 
        on different recognition tasks, showing that the compression does not 
        alter the effectiveness and generality of the learned representations.
\end{enumerate}

%% file: related.tex
\section{Related Work}
Redundancy in parameterization of neural networks is a well-known phenomenon. 
Indeed, Denil~\etal showed that it is possible to predict the $~95\%$ of the 
weights of a neural network (without drop in accuracy) just using the remaining 
$5\%$ of the weights \cite{denil2013predicting}.

In literature, many approaches have been proposed to deal with the task of 
reducing the size of the networks without affecting performance, to improve both 
computational and memory efficiency. One of the first explored ideas is 
\emph{weight sharing}, \ie to constrain some of the weights of a layer in a 
neural network to be the same. Among them, we recall
\begin{enumerate*}[label=(\roman*)]
    \item the use of \emph{locally connected features} 
        \cite{coates2011analysis};
    \item \emph{tiled convolutional networks} \cite{gregor2010emergence};
    \item \emph{convolutional neural networks} (\muacro{CNN}s) 
        \cite{lecun1998gradient}. 
\end{enumerate*}
Based on the same idea, \emph{HashedNet}s exploit a hash function to randomly 
group connection weights, so that all connections within the same hash bucket 
share a single weight value \cite{chen2015compressing}. 

Another interesting approach is to take an existing network model and compress 
it in a \emph{lossy} fashion. A fairly straightforward approach proposed by 
Denton~\etal employs singular value decomposition to a pre-trained \muacro{CNN} 
model, so to get a low-rank approximation of the weights while keeping the 
accuracy within 1\% of the original model \cite{denton2014exploiting}. Another 
approach to lossy compression is \emph{network pruning}. This technique tries to 
remove edges in a neural architecture with small weight magnitudes. Viable 
implementations of network pruning are: 
\begin{enumerate*}[label=(\roman*)]
    \item \emph{weight decay} \cite{hanson1989comparing};
    \item \emph{Optimal Brain Damage} \cite{Cun:1990:OBD:109230.109298}; 
    \item \emph{Optimal Brain Surgeon} \cite{hassibi1993optimal}.
\end{enumerate*}
\emph{Optimal Brain Damage} and \emph{Optimal Brain Surgeon} prune networks 
based on the Hessian of the loss function and the results obtained suggest that 
such pruning is more accurate than magnitude-based pruning and weight decay. 
Recently, Han~\etal achieved in \cite{han2015learning} pruned networks by 
setting to zero the weights below a threshold, without drop of accuracy and 
reducing the final number of weights by an order of magnitude. More recently, 
Han~\etal extended in \cite{Han2015} the aforementioned approach: they 
lengthened the compression pipeline by \emph{quantizing} the network weights (to 
8 bits or less) and finally Huffman encoding is employed. 
They also showed that pruning and quantization are able to compress the network 
without interfering each other.
This technique, called 
\emph{Deep Compression}, has been deployed on custom hardware accelerator called 
Efficient Inference Engine, achieving substantial speedups and energy savings 
\cite{Han:2016:EEI:3001136.3001163}.

\emph{Deep Compression} wasn't the first technique exploiting quantization to 
achieve network compression. Indeed, quantization approaches have been largely 
explored, since it is well-known that deep networks are not highly sensitive to 
floating point precision. In \cite{gong2014compressing}, for the first time, 
Gong~\etal employed quantization techniques for deep architectures, achieving a 
compression rate of 4-8$\times$ just using quantization, while keeping 
the accuracy loss within 1\% on the \ilsvrc dataset. In \cite{hwang2014fixed}, 
Hwang~and~Sung proposed an optimization method for the fixed-point networks with 
ternary weights and 3-bit activation functions, while Vanhoucke~\etal explored in 
\cite{vanhoucke2011improving} a fixed-point implementation with 8-bit integer 
activation functions (vs 32-bit floating point).

The extreme version of weight quantization is to build a network with 
only \emph{binary weights}. Courbariaux~\etal presented \emph{BinaryConnect} for 
training a network with +1/-1 weights \cite{courbariaux2015binaryconnect}, and 
Hubara~\etal introduced \emph{BinaryNet} for training a network with both binary 
weights and binary activation functions \cite{NIPS2016_6573}. Both \emph{BinaryConnect} 
and \emph{BinaryNet} achieve good performance on small datasets, but they perform 
worse than their full-precision counterparts by a wide margin on large-scale 
datasets. In \cite{rastegari2016xnor}, Rastegari~\etal presented \emph{Binary 
Weight Network}s and \emph{XNOR-Net}s, two approximations to standard 
\muacro{CNN}s that are shown to highly outperform BinaryConnect and BinaryNet on 
ImageNet.

Our work is inspired by \emph{Deep Compression} \cite{Han2015}, and it enhances 
the pruning stage of the compression pipeline by making it differentiable with 
respect to the threshold weights.

On the other hand, transfer learning in the field of machine learning is the 
ability to exploit the knowledge gained while solving one specific problem and 
applying it to a different related problem \cite{michalski1983theory}. Many 
approaches have been proposed \cite{thrun2012learning}, and it has been 
demonstrated that deep neural networks have very good transfer learning 
capabilities \cite{yosinski2014transferable} and that can be applied to a very 
different set of related problems outperforming methods specifically designed to 
solve them \cite{sharif2014cnn}. Considering the aforementioned results, in this 
work we tried to assess if the representations learned using a compressed 
network can achieve comparable results compared to those obtained by the 
non-compressed one.

%% file: method.tex
\section{Method}
Given the structure of a generic neural network $\Network{N}$---the number of 
nodes, their connections, the activation functions employed, and the untrained 
weights---the complete pipeline to compress $\Network{N}$ is composed of three 
parts:
\begin{enumerate*}[label=(\roman*)]
    \item building a \emph{sibling network} $\Network{N}_{\text{s}}$ that is 
        explicitly able to shrink the trainable weights of $\Network{N}$;
    \item training $\Network{N}_{\text{s}}$ by means of a gradient descent-based 
        technique, where a regularization term $\mathcal{L}_t$ is introduced to 
        enforce the shrinkage of the weights;
    \item building 
        $\Network{N}_{\text{p}}$ from $\Network{N}_{\text{s}}$, where 
        $\Network{N}_{\text{p}}$ is the \emph{pruned version} of the original 
        network $\Network{N}$.
\end{enumerate*}

To better describe the aforementioned steps, in the next section we will 
present some useful definitions.

\subsection{Preliminaries}
$\Network{N}$ can be seen as the functional composition of many layer functions 
$\Layer{L}_i$, where the $i$ subscript determines the depth of the layer within 
the network: $\Network{N} = \Layer{L}_d \circ \ldots \circ \Layer{L}_1$. Each 
layer function is identified by: 
\begin{enumerate}[label=(\roman*),noitemsep,topsep=0pt]
    \item the parameterized family of functions $\mathcal{T}$ where the layer 
        function belongs;
    \item the collection of (learnable) weights $\Set{W} \coloneqq 
        \{\ldots, \Matrix{W}_j, \ldots\}$ needed to specify the layer function within 
        $\mathcal{T}$.
\end{enumerate}
For example, if we want to describe a \emph{fully-connected} layer with activation $f$, 
$\mathcal{T}$ is the family of affine transformations followed by $f$, and 
$\Set{W}$ is the collection of weights and bias of the affine map.
Hereinafter, we will write $L[\Set{W}_i]$ when the type of layer is apparent 
from the context, in order to stress the weight dependency and simplifying the 
notation. Moreover, given a function $f$ and a set $\Set{A}$, we will use the 
convention that $f(\Set{A})$ represents the image of the function $f$ when 
applied on the elements of the set $\Set{A}$, \ie $f(\Set{A}) \coloneqq \{ f(x) 
\mid x \in \Set{A} \}$.

\begin{figure}[t]
\centering
    \begin{subfigure}[t]{.70\columnwidth}
    \includegraphics[width=\columnwidth]{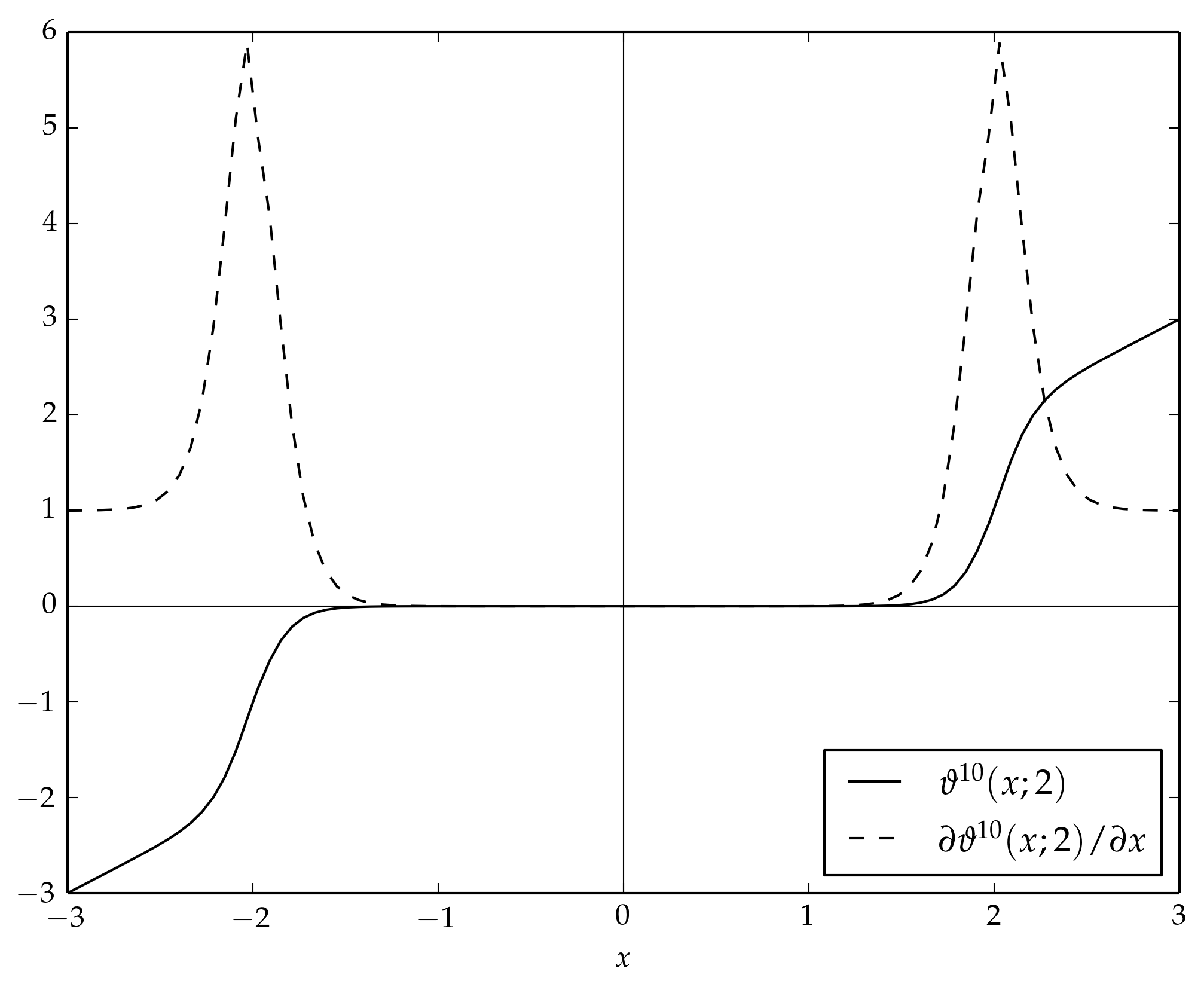}
    \caption{The solid line represents the function $\vartheta^{10}(x; 2)$. The 
        dashed line represents its derivative with respect to $x$.}
    \end{subfigure}
    \begin{subfigure}[t]{.70\columnwidth}
    \includegraphics[width=\columnwidth]{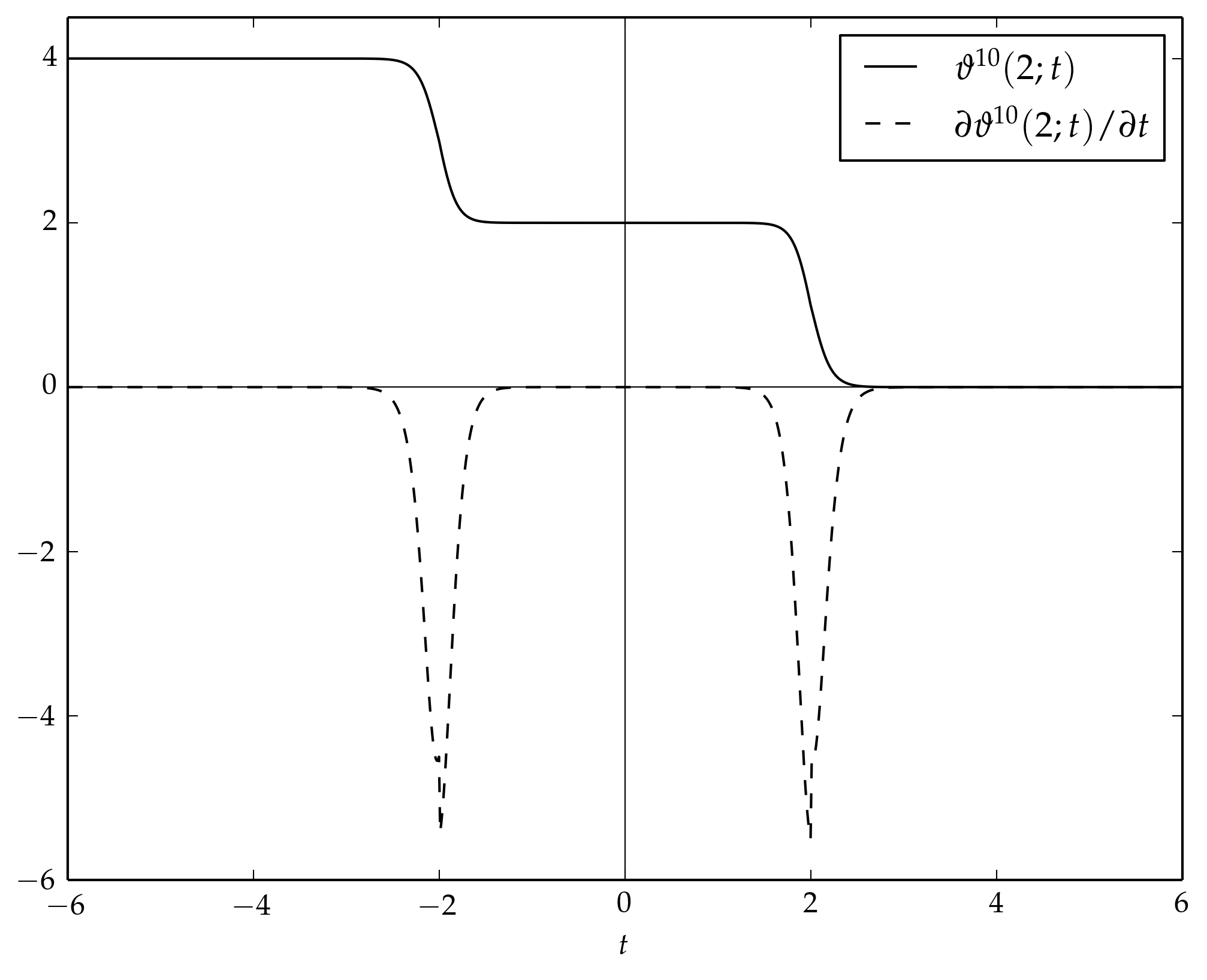}
    \caption{The solid line represents the function $\vartheta^{10}(2; t)$. The 
        dashed line represents its derivative with respect to $t$.}
    \end{subfigure}
    \caption{\label{fig:pruning-function-continuous}Representation of the 
function $\vartheta^\alpha$ of Equation~\eqref{eq:pruning-function}.}
\vspace{-2em}
\end{figure}
\begin{definition}[Pruning function]
The \emph{pruning function} $\vartheta^\alpha(\cdot; t)$, 
needed for the compression procedure, is defined as follows:
\begin{equation}\label{eq:pruning-function}
    \begin{aligned}
        \vartheta^\alpha(x; t) \coloneqq  & 
        \hphantom{-.}\relu(x-t) + t \cdot \sigma(\alpha (x-t)) + \\
        & - \relu(-x-t) - t \cdot \sigma(\alpha(-x-t)),
    \end{aligned}
\end{equation}
where $\alpha$ and $t$ are positive real numbers, $\sigma$ is the 
sigmoidal function $\sigma(x) \coloneqq {(1 + 
\mathrm{e}^{\Minus x})}^{\Minus 1}$, and $\relu(x) \coloneqq \max(0, x)$ is 
the \emph{Rectified Linear Unit} function (see 
Figure~\ref{fig:pruning-function-continuous}).
\end{definition}
Note that, the purpose of the pruning function is to force \emph{toward zero} all 
the elements of the domain within $(-t - \Delta, +t + \Delta)$ (for a suitable 
$\Reals \ni \Delta \geq 0$); the variable $t$ acts then as a threshold 
variable. On the other hand, the parameter $\alpha$ determines the speed how 
quickly the pruning takes place within such interval. 

We are now showing that $\vartheta^\alpha$ is learnable with respect to the variables $x$ and $t$.
The (weak) partial derivatives of 
$\vartheta^\alpha$ with respect of $x$ and $t$ are given by:
\begin{align}
    \frac{\partial \vartheta^\alpha}{\partial x} = & H(x-t) + H(-x-t) + \nonumber\\
    & + \alpha t \cdot \sigma(\alpha(x-t)) [1 - \sigma(\alpha(x-t))] +\nonumber\\ 
    & + \alpha t \cdot \sigma(\alpha(-x-t)) [1 - \sigma(\alpha(-x-t))],\label{eq:pruning-function-derivative-alpha}\\
    \frac{\partial \vartheta^\alpha}{\partial t} = & - H(x-t) + H(-x-t) + \nonumber\\
    & + \sigma(\alpha(x-t)) - \sigma(\alpha(-x-t)) + \nonumber\\
    & - \alpha t \cdot \sigma(\alpha(x-t)) [1 - \sigma(\alpha(x-t))] + \nonumber\\
    & + \alpha t \cdot \sigma(\alpha(-x-t)) [1 - \sigma(\alpha(-x-t))]\label{eq:pruning-function-derivative-t},
\end{align}
where $H(x)$ is the \emph{Heaviside step function}: $H(x) = 1$ if $x\geq0$, $H(x) = 0$ otherwise. 
Equations~\eqref{eq:pruning-function-derivative-alpha} and 
\eqref{eq:pruning-function-derivative-t} show that the (weak) partial 
derivatives of $\vartheta^\alpha(x; t)$ are different than zero almost everywhere, 
thus allowing the parameter $t$ to be learnable by means of gradient 
descent (see also 
Figure~\ref{fig:pruning-function-continuous}).

\begin{figure}[t]
\centering
    \begin{subfigure}[t]{.70\columnwidth}
        \includegraphics[width=\columnwidth]{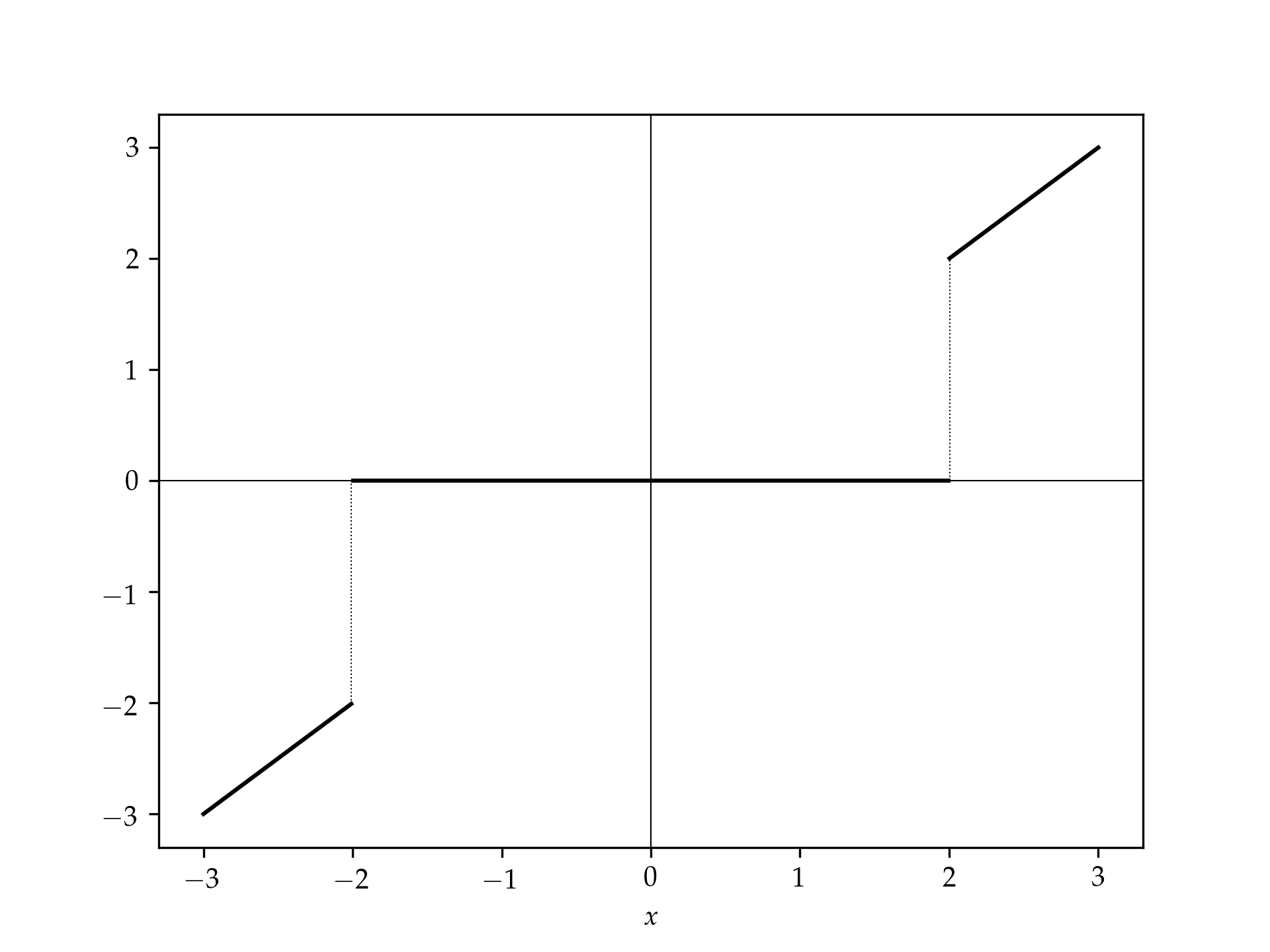}
        \caption{The function $\bar\vartheta(x; 2)$.}
    \end{subfigure}
    \begin{subfigure}[t]{.70\columnwidth}
        \includegraphics[width=\columnwidth]{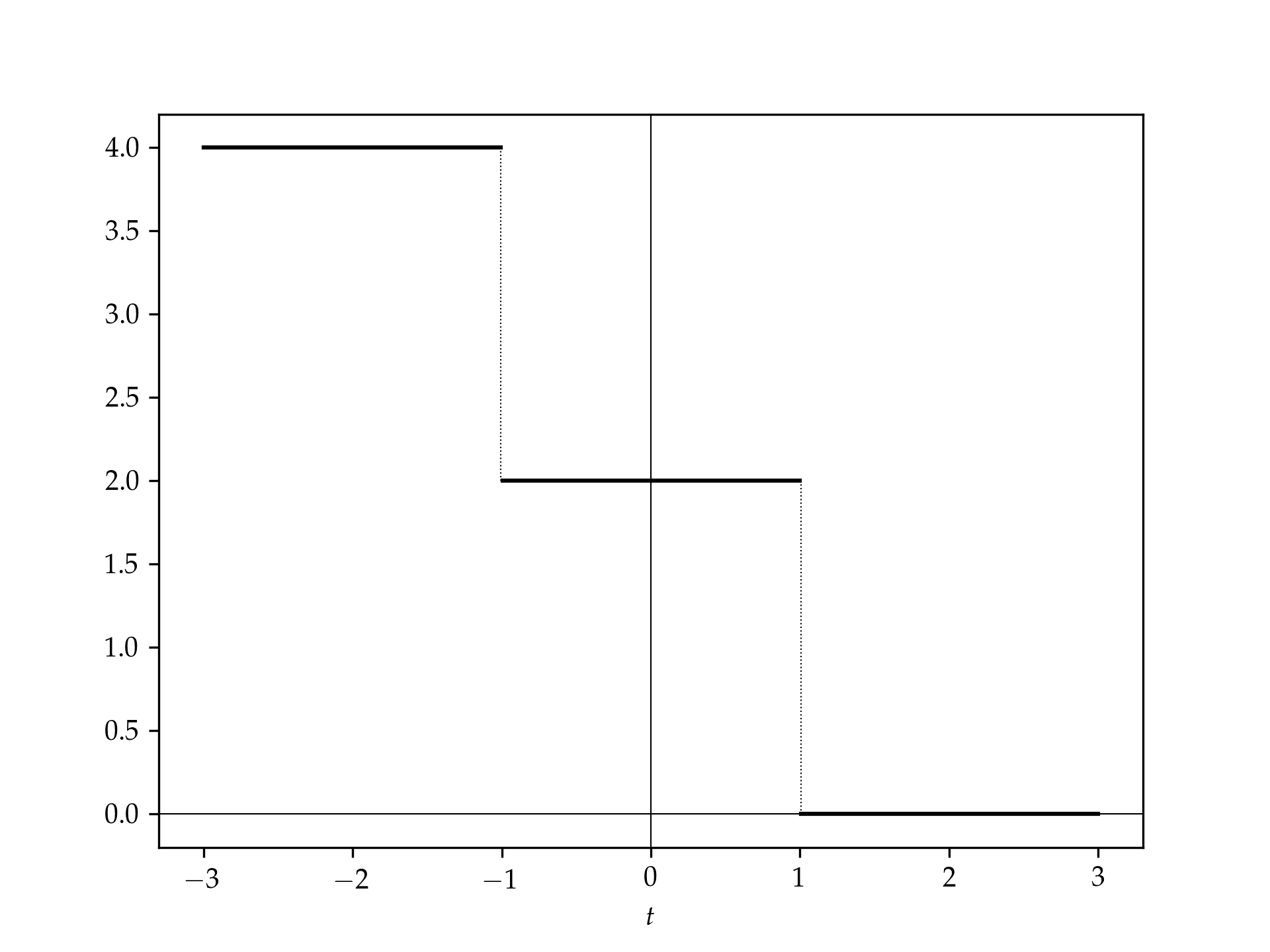}
        \caption{The function $\bar\vartheta(2; t)$.}
    \end{subfigure}    
    \caption{\label{fig:pruning-function-hard}The graph of the function 
$\bar\vartheta(x; t)$ of Equation~\eqref{eq:pruning-function-hard}. Note how the 
derivative with respect of $t$ is zero almost everywhere.
    }
\vspace{-2em}
\end{figure}
It is interesting to notice that, $\vartheta^\alpha(\,\cdot\,;t)$ is a smoothed 
version of the \emph{thresholded linear} function $\bar\vartheta(\cdot; t)$.

\begin{definition}[Thresholded linear function]
The thresholded linear function $\bar\vartheta(\cdot; t)$ is defined as follow 
\cite{rozell2008sparse, konda2014zero}:
\begin{align}\label{eq:pruning-function-hard}
    \bar\vartheta(x; t) \coloneqq & \hphantom{-.} \relu(x-t) + t 
    \cdot H(x-t) + \phantom{.}\nonumber\\
    &- \relu(-x-t) - t \cdot H(-x-t).
\end{align}
\end{definition}
Note that, $\vartheta^\alpha(x; t)$ converges 
weakly to $\bar\vartheta(x; t)$ when $\alpha \to 
\infty$. Moreover, the weak partial derivatives of $\bar\vartheta(x; t)$ with 
respect to $x$ and $t$ are:
\begin{align*}
    & \frac{\partial \bar\vartheta}{\partial x} = H(x-t) + 
        t\cdot\delta(x-t) + H(-x-t) + t\cdot\delta(x+t),\\
    & \frac{\partial \bar\vartheta}{\partial t} = -t\cdot\delta(x-t) + 
        t\cdot\delta(x+t),
\end{align*}
where $\delta$ is the \emph{Dirac's delta}, thus forbidding any sort of learning 
procedure on the variable $t$ (see 
Figure~\ref{fig:pruning-function-hard})\footnote{$\partial \bar\vartheta / 
\partial t$ is almost everywhere equal to zero.}.

\subsection{Sibling networks}
\subsubsection{Building}
Given a network $\Network{N} = L[\Set{W}_d] \circ \ldots \circ L[\Set{W}_1]$, the sibling 
network $\Network{N}_{\text{s}}$ is defined as follows:
\begin{equation}\label{eq:sibling-network}
    \Network{N}_\text{s} \coloneqq \Layer{L}[\vartheta^{\alpha}(\Set{W}_d; t_d)] \circ \ldots 
        \circ \Layer{L}[\vartheta^{\alpha}(\Set{W}_1; t_1)].
\end{equation}
Note that, in the equations for the networks $\Network{N}$ and 
$\Network{N}_\text{s}$ we did not specify the type of each layer so to reduce 
the clutter in the notation. Moreover, in the previous equation $\alpha$ is an 
hyper-parameter, while all the $t_i$ are learnable weights. For the sake of 
the clarity, we are assuming one threshold variable per layer, as well as 
that there is one $\alpha$ hyper-parameter shared by all the layers. These
simplifying assumptions can be easily relaxed.

\subsubsection{Training}\label{sec:sibling-network-training}
The learning of all the $\Matrix{W}_j$ and $t_j$ of the sibling network $\Network{N}_{\text{s}}$ is performed by 
means of gradient descent. The loss function to be minimized is given by 
$\mathcal{L} \coloneqq \mathcal{L}_0 + \mathcal{L}_{\text{wd}} + \mathcal{L}_t$, 
with $\mathcal{L}_0$ the basic loss function for the problem under analysis (\eg 
cross-entropy), where: 
\begin{align}
    & \mathcal{L}_{\text{wd}} \coloneqq \lambda_{\text{wd}} \sum_i 
        \sum_{\Matrix{W}_j \in \Set{W}_i} \Norm{\Matrix{W}_j}_2^2, \nonumber\\
    & \mathcal{L}_{t} \coloneqq \lambda_t \sum_i \sum_{j \mid 
        \Matrix{W}_j\in\Set{W}_i } \Norm{\vartheta^\alpha(\Matrix{W}_j; t_j)}_1, 
    \label{eq:regularizer-pruning}
\end{align}
with $\Norm{\Matrix{X}}_p$ the entry-wise $p$-norm of the matrix $\Matrix{X}$.
Namely,
\begin{enumerate}[label=(\roman*),noitemsep,topsep=0pt]
    \item $\mathcal{L}_{\text{wd}}$ is the usual weight decay regularizer 
        on all the $\Matrix{W}_j$ variables, with $\lambda_\text{wd}$ the 
        corresponding hyper-parameter;
    \item $\mathcal{L}_t$ is a regularizer used to speed-up pruning, where 
        $\lambda_t$ is its hyper-parameter. 
\end{enumerate}
The optimization of the training function is performed using one variant of 
stochastic gradient descent, with the following caveats:
\begin{enumerate}[label=(\roman*),noitemsep,topsep=0pt]
    \item the regularizer $\mathcal{L}_t$ is optimized only with respect to all 
        the variables $t_j$, \ie the contribution to the gradient given by 
        $\mathcal{L}_t$ with respect to all the $\Matrix{W}_j$ is zero;
    \item the learning rate of all the variables $t_j$ is 
        slowed by a factor $\rho$ so to reduce the effective 
        learning rate of the thresholds with respect to the other parameters;
    \item the learning of all $t_j$ is performed by enforcing 
        non-negativity;
    \item weights are initialized randomly as in a customary neural network (\eg 
        Glorot initialization \cite{glorot2010understanding}), while initial 
        values of the thresholds are set such that a small amount $p$ 
        (\eg $0\%\div10\%$) of weights are effectively below the threshold at 
        the beginning of the training.
\end{enumerate}
The purpose of the regularizer $\mathcal{L}_t$ is to enforce sparsity of the 
weights under the mapping of the pruning function $\vartheta^\alpha$. Since it 
affects only the learning of all the thresholds $t_j$ (\ie it is not involved 
in the partial derivative of the loss with respect to any $\vect{W}_j$), it 
effectively moves the pruning thresholds so to increase the \emph{after-mapping 
sparsity}. Instead, the weight decay regularizer $\mathcal{L}_{\text{wd}}$ is used to 
enforce the weights to gather around all the threshold values $t_j$ (see 
Figure~\ref{fig:lenet300100-learned-weights} in Section~\ref{sec:results-pruning-lenet} for a comparison between the 
distribution of the learned weights of the first layer of a \lenetff on \mnist 
with and without the $\mathcal{L}_{\text{wd}}$ regularizer).

\subsubsection{Pruning}
The sibling networks $\Network{N}_{\text{s}}$ learned accordingly to 
Section~\ref{sec:sibling-network-training} has few zero weights, due to the 
smooth behavior of the function $\vartheta^\alpha$ that we used to enforce 
weights shrinking. However, thanks to the $\vartheta^\alpha$ itself, many of the 
weights of the network $\Network{N}_\text{s}$ are \emph{near} zero. For this 
reason, we get an actual pruned network $\Network{N}_{\text{p}}$ from 
$\Network{N}_{\text{s}}$ by setting to zero all the learned weights of 
$\Network{N}_{\text{s}}$ that under the mapping of $\theta^\alpha$ are below a 
certain (small) cutoff value $\gamma$, which has to be considered as a 
hyper-parameter. We found out empirically that $\gamma \sim 
10^{\Minus 3}$ is a good starting choice in almost all the cases.

Precisely, denoting with $\widetilde{\Set{W}}_j$ the weights learned during 
the training phase of the sibling network and with $\tilde t_j$ the learned 
threshold parameters, the pruned network $\Network{N}_{\text{p}}$ is given by:
\begin{align}\label{eq:pruned-network}
    \Network{N}_{\text{p}} \coloneqq & \Layer{L}[\vartheta^{\alpha}_{\text{inv}}(
        \bar\vartheta(\vartheta^{\alpha}(\widetilde{\Set{W}}_d; \tilde{t}_d); 
        \gamma); \tilde{t}_d)] 
        \circ \\ 
        & \ldots \circ \Layer{L}[\vartheta^{\alpha}_{\text{inv}}(
        \bar\vartheta(\vartheta^{\alpha}(\widetilde{\Set{W}}_1;
            \tilde{t}_1); \gamma); 
        \tilde{t}_1)],\nonumber
\end{align}
where $\vartheta^{\alpha}_{\text{inv}}(x; t)$ is the inverse function of 
$\vartheta^{\alpha}(x; t)$ with respect to $x$ when $t$ is given. Note that 
$\vartheta^{\alpha}_{\text{inv}}$ exists, since $\vartheta^{\alpha}$ is 
bijective. Moreover, since $\vartheta^{\alpha}(0; t) = 0$ we have that 
$\vartheta^{\alpha}_{\text{inv}}(0; t) = 0$.

%% file: results.tex
\section{Results}\label{sec:results}
The experiments we performed can be split in two groups:
\begin{enumerate}[label=(\roman*),noitemsep,topsep=0pt]
    \item we pruned \lenetff and \lenetconv on \mnist dataset, as well as
        \alexnet on \ilsvrc dataset; for these networks we assessed the 
        compression ratio due to pruning as well as the drop in accuracy against 
        their non-pruned counterparts (see 
        Section~\ref{sec:results-pruning});
    \item for the pruned \alexnet on \ilsvrc we studied the generalization 
        performance in the transfer learning scenario considering different 
        recognition tasks, such as object image classification, scene 
        recognition, fine grained recognition, attribute detection, and 
        image retrieval (see Section~\ref{sec:results-tl}).
\end{enumerate}

\subsection{Pruning}\label{sec:results-pruning}
We pruned \lenetff and \lenetconv on \mnist dataset, and \alexnet 
on \ilsvrc dataset. 

\mnist is a large database of gray-scale handwritten digits, made of 60K training 
images and 10K testing images \cite{mnist}. On the other hand, the \ilsvrc 
dataset is a 1000 classes classification task with 1.2M training examples and 
50k validation examples.

All our networks were built and trained using the Keras 
framework \cite{chollet2015keras} on top of TensorFlow 
\cite{tensorflow2015-whitepaper}.
The size of the networks and their accuracy before and after pruning are shown 
in Table~\ref{tab:results-macro}, together with the performance achieved by 
Han~\etal in \cite{han2015learning}, the paper of \emph{Deep Compression} 
focusing on the pruning stage of the compression pipeline. The technique 
presented in this paper keeps the error rate of the pruned networks comparable 
to the non-pruned counterparts as in the current state-of-the-art, while 
achieving better pruning rate. In the experiments we performed, our pruning 
technique saved network storage by $12\times$ to $19\times$ across different 
networks, thus increasing Han~\etal compression rate by a factor 
$\approx\!1.3\times$ to $\approx\!1.6\times$ and reducing by $14\%$ to $33\%$ 
the number of weights retained by \emph{Deep Compression}.

\begin{figure}%[t]
\centering
    \includegraphics[width=.7\columnwidth]{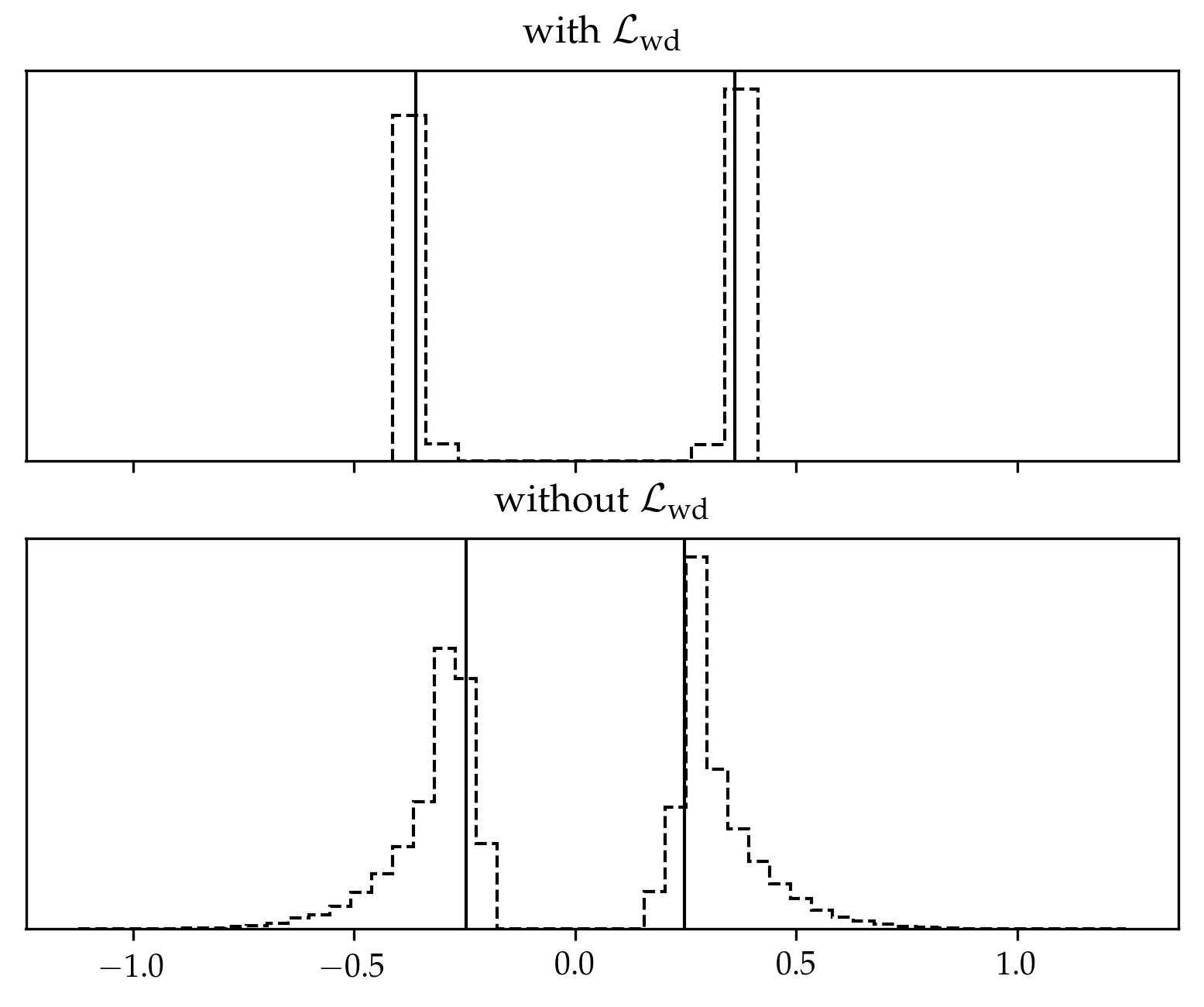}
    \caption{\label{fig:lenet300100-learned-weights}The figure shows the 
distribution of the trained weights of the first layer of two pruned \lenetff on 
\mnist, with $\alpha=100$ and a cutoff $\gamma=10^{\Minus 3}$. The vertical 
solid lines represent the learned thresholds. The top plot was trained with 
weight decay regularizer, while the bottom one without it. Note how in the first 
case the weights group around the threshold values.}
\vspace{-2em}
\end{figure}

\subsubsection{\lenetff and \lenetconv on \mnist}\label{sec:results-pruning-lenet}
We first experimented on \mnist dataset with \lenetff and \lenetconv networks 
\cite{lecun1998gradient}. \lenetff is a fully connected network made of two 
hidden layers, with 300 and 100 neurons respectively, achieving 
$1\%\!\div\!2\%$ error rate on \mnist.

\lenetconv is a convolutional network that has two convolutional layers 
and two fully connected layers, achieving $<1\%$ error rate on \mnist.
All the convolutional layers of \lenetconv have been pruned by learning a 
different threshold weight \emph{per filter}, thus partially relaxing the 
simplifying assumption we used to write Equation~\eqref{eq:sibling-network}.

All the networks were trained using Adam optimizer \cite{adam} with learning 
rate $10^{\Minus3}$. In addition, pruned networks were learned with the 
following configuration of the hyper-parameters: $\alpha=10^2$, 
$p=10^{\Minus1}$, $\rho=\lambda_t = 10^{\Minus2}$, $\gamma=10^{\Minus3}$, 
$\lambda_{\text{wd}}=10^{\Minus4}$.

Table~\ref{tab:results-macro} shows that these networks on \mnist can be pruned 
with basically no drop in accuracy with the respect of their non-compressed 
counterparts. The pruning procedure achieves a storage saving of $15\!\times$ 
and $19\!\times$, reducing by 33\% (7K) and 14\% (14K) the number of weights 
retained by \emph{Deep Compression}, for \lenetff and \lenetconv respectively. 

Table~\ref{tab:rate-layer-everything} shows the per-layer statistics of the 
pruning procedure. It is interesting to notice that, our approach converged to a 
network configuration where the pruning ratio is significantly higher for the 
biggest layers.

Figure~\ref{fig:lenet300100-learned-weights} shows the comparison between 
the distribution of the first layer weights of a \lenetff with and without 
weight-decay. It is possible to notice that, the weights are much more gathered 
around the learned threshold when the weight-decay parameter is used during the 
learning process. We speculate such a grouping might make weights quantization 
in \emph{Deep Compression} \cite{Han2015} more efficient.

\subsubsection{\alexnet on \ilsvrc}
We implemented the original \alexnet model \cite{krizhevsky2012imagenet} from 
scratch using the Keras framework. The training of the non-pruned and pruned 
\alexnet were performed using stochastic gradient descent with starting learning 
rate $10^{\Minus2}$, step-wise learning rate decay, learning rate multiplier 
equal to $2$ for all the bias weights, weight-decay hyper-parameter 
$\lambda_\text{wd} = 5\cdot10^{\Minus4}$, and momentum $0.9$, exactly as in the 
original paper \cite{krizhevsky2012imagenet}. On the other hand, the compression 
procedure was carried out with the hyper-parameters $\alpha=10^2$, $p=0$, 
$\rho=10^{\Minus11}$, $\lambda_t = 10^{\Minus2}$, $\gamma=10^{\Minus3}$.

\begin{table}
    \begin{center}
\resizebox{\columnwidth}{!}{
    \begin{tabular}{l|cccccc}
        \hline
        \hline
        Network & $\Delta_1$ Err. & $\Delta_5$ Err. & Weights & Pruning\\
        \hline
        \hline
        \lenetff reference & ---  & --- & 266K & --- \\
        \lenetff pruned (Han~\etal) & $\approx\!-0.1\%$ & --- & 21K & $12\times$ \\
        \lenetff pruned (this paper) & $\approx\!+ 0.1\%$ & --- & $\bm{14\text{K}}$ & $\bm{19\times}$ \\
        \hline
        \lenetconv reference & --- & --- & 431K & --- \\
        \lenetconv pruned (Han~\etal) & $\approx\!- 0.1\%$ & --- & 34K & $12\times$ \\
        \lenetconv pruned (this paper) & $\approx\!- 0.0\%$ & --- & $\bm{29\text{K}}$ & $\bm{15\times}$ \\
        \hline
        \alexnet reference & --- & --- & 61M & --- \\
        \alexnet pruned (Han~\etal) & $\approx\!- 0.0\%$ & $\approx\!- 0.0\%$ & 7M & $9\times$ \\
        \alexnet pruned (this paper) & $\approx\!+ 1.1\%$ & $\approx\!+ 1.0\%$ & $\bm{5\text{M}}$ & $\bm{12\times}$ \\
        \hline
        \hline
    \end{tabular}
}
    \end{center}
\caption{\label{tab:results-macro}The table shows the difference in Top-1 
($\Delta_1$) and Top-5 ($\Delta_5$) errors between the networks pruned in this 
paper and their reference non-pruned implementation. It also shows the pruning 
performance. Similarly, it is also shown $\Delta_1$/$\Delta_5$ and pruning rate 
achieved by Han~\etal in \cite{han2015learning}.}
\end{table}

\begin{table}
    \begin{center}
\resizebox{\columnwidth}{!}{
    \begin{tabular}{l|cccccccc|c}
        \hline
        \hline
        &\multicolumn{9}{c}{\lenetff} \\
        \hline
        & \multicolumn{3}{c}{\muacro{fc1}} & \multicolumn{3}{c}{\muacro{fc2}} & \multicolumn{2}{c|}{\muacro{fc3}} & total\\
        Weights & \multicolumn{3}{c}{235K} & \multicolumn{3}{c}{30K} & \multicolumn{2}{c|}{1K} &  431K\\
        Pruning & \multicolumn{3}{c}{94.7\%} & \multicolumn{3}{c}{83.8\%} & \multicolumn{2}{c|}{11.9\%} & 94.7\%\\
         & \multicolumn{3}{c}{19$\times$} & \multicolumn{3}{c}{6$\times$} & \multicolumn{2}{c|}{1$\times$} & 19$\times$\\
        \hline
        \hline
        &\multicolumn{9}{c}{\lenetconv} \\
        \hline
        & \multicolumn{2}{c}{\muacro{conv1}} & \multicolumn{2}{c}{\muacro{conv2}} & \multicolumn{2}{c}{\muacro{fc3}} & \multicolumn{2}{c|}{\muacro{fc4}} & total\\
%         Weights & 0.5K & 25K & 400K & 1K & 431K\\
        Weights & \multicolumn{2}{c}{0.5K} & \multicolumn{2}{c}{25K} & \multicolumn{2}{c}{400K} & \multicolumn{2}{c|}{1K} &  431K\\
        Pruning & \multicolumn{2}{c}{31.1\%} & \multicolumn{2}{c}{82.2\%} & \multicolumn{2}{c}{96.6\%} & \multicolumn{2}{c|}{41.7\%} &  93.3\%\\
         & \multicolumn{2}{c}{1$\times$} & \multicolumn{2}{c}{6$\times$} & \multicolumn{2}{c}{29$\times$} & \multicolumn{2}{c|}{2$\times$} &  15$\times$\\
%         Pruning (\%) & 31.1\% & 82.2\% & 96.6\% & 41.7\% & 93.3\% (15$\times$)\\
        \hline
        \hline
        &\multicolumn{9}{c}{\alexnet} \\
        \hline
        & \muacro{conv1} & \muacro{conv2} & \muacro{conv3} & \muacro{conv4} & \muacro{conv5} & \muacro{fc6} & \muacro{fc7} & \muacro{fc8} & tot\\
        Weights & 35K & 307K & 885K & 663K & 442K & 38M & 17M & 4M & 61M\\
        Pruning & 3.8\% & 34.9\% & 24.3\% & 36.7\% & 33.5\% & 96.8\% & 91.5\% & 78.5\% & 91.5\%\\
         & $1\times$ & $2\times$ & $1\times$ & $2\times$ & $2\times$ & $31\times$ & $12\times$ & $5\times$ & $12\times$\\
        \hline
        \hline
    \end{tabular}
}
    \end{center}
\caption{\label{tab:rate-layer-everything}The table shows how the pruning rate 
behaves for each layer of \lenetff and \lenetconv trained on \mnist, and 
\alexnet trained on \ilsvrc.}
\end{table}

Table~\ref{tab:results-macro} shows that our procedure can achieve a memory 
saving of about $12\!\times$, with a small drop in the accuracy compared to the
non-compressed counterpart (\ie $\approx\!1.1\%$ top-1 and 
$\approx\!1.0\%$ top-5). Such a pruning performance translates into a reduction 
of 2M (29\%) weights compared to the number of weights retained by 
\emph{Deep Compression}.

Table~\ref{tab:rate-layer-everything} shows the per-layer statistics of the 
pruning procedure. Again, our approach converged to a network compression where 
the pruning ratio is significantly higher for the biggest layers of the 
networks, with a staggering $31\!\times$ for the biggest layer of the network, 
\ie \muacro{fc6}.

%% file: cnnfeatures.tex
\begin{table*}
\centering
\resizebox{0.80\textwidth}{!}{
\begin{tabular}{lllccr}
\hline
\hline
Task & Dataset & Performance Measure & Uncompressed & Compressed & Difference\\
\hline
Image classification & Pascal VOC 2007 \cite{everingham2011pascal}           	& mean Average Precision (mAP) & 0.6235 & 0.6132 & -0.0103\\
					 & MIT-67 indoor scenes \cite{quattoni2009recognizing}    	& Accuracy & 0.5440 & 0.5425 & -0.0015\\
\hline
Fine grained recognition & Birds (CUB) 200-2011 \cite{wah2011caltech}	& Accuracy & 0.5043 & 0.5173 &  0.0130\\
					 & Oxford 102 flowers \cite{nilsback2008automated}   		& Accuracy & 0.8477 & 0.8542 &  0.0065\\
\hline					 
Attribute detection  & UIUC 64 objects attributes \cite{farhadi2009describing} 	& mean Area Under Curve (mAUC) & 0.7999 & 0.7953 & -0.0046\\
					 & H3D person attributes \cite{bourdev2011describing}		& mean Average Precision (mAP) & 0.5664 & 0.5646 & -0.0018\\
\hline
Visual instance retrieval &	Oxford5k buildings \cite{philbin2007object} & mean Average Precision (mAP) & 0.3471 & 0.3901& 0.0430\\
						  &	Paris6k buildings \cite{philbin2008lost} & mean Average Precision (mAP) & 0.5958& 0.6007& 0.0049\\
						  &	Sculptures6k \cite{arandjelovic2011smooth}      & mean Average Precision (mAP)& 0.3093& 0.2982& -0.0111\\
						  &	Holidays dataset \cite{jegou2008hamming}	       & mean Average Precision (mAP)& 0.7302& 0.7187& -0.0115\\
						  &	UKbench	\cite{nister2006scalable}       &  Recall@4& 0.8770& 0.8904& 0.0134\\
\hline
\hline
\end{tabular}
}
\caption{\label{tab:cnnfeatures}Transfer learning performance of the features 
extracted from the \alexnet network trained and compressed on ImageNet \ilsvrc 
compared to those extracted from the same uncompressed network. The comparison 
is performed on different recognition tasks: object image classification, scene 
recognition, fine grained recognition, attribute detection and image retrieval 
applied to a diverse set of datasets.}
\vspace{-2em}
\end{table*}

\subsection{Transfer Learning}\label{sec:results-tl}
In the previous section we showed that we are able to heavily compress a 
deep neural network keeping the same recognition performance on the training 
dataset. Since \muacro{CNN} are very often used for transfer learning as feature 
extractors due to the the effectiveness and generality of the learned 
representations \cite{sharif2014cnn}, in this section we investigate how the 
compressed features perform on various recognition tasks and different datasets. 
We used features extracted from the \alexnet network trained and compressed 
on \ilsvrc dataset as a generic image representation to tackle the diverse range 
of recognition tasks tested in \cite{sharif2014cnn}, \ie: object image 
classification, scene recognition, fine grained recognition, attribute detection 
and image retrieval applied to many datasets.

For all the experiments we resized the input image to $227 \times 227$ and we used 
the last fully connected layer (\ie layer \muacro{fc7}) of the network as our 
feature vector. This gives a vector of 4096 dimension that is further $L2$ 
normalized to unit length for all the experiments. For all the different 
classification and recognition tasks considered we used the 4096 dimensional 
feature vector in combination with a linear Support Vector Machine 
\cite{cortes1995support, fan2008liblinear}. For visual instance retrieval task 
we adopted the Euclidean distance to compute the visual similarity between a query 
and the images from target dataset. 

% \vspace{-.2cm}
\subsubsection{Image Classification}\label{sec:image_classification}
The first problem faced was image classification of objects and scenes. The task 
is to assign (potentially multiple) semantic labels to an image. Two datasets 
were considered for two different recognition tasks: the Pascal VOC 2007 for 
object image classification \cite{everingham2011pascal} and the MIT-67 indoor 
scenes \cite{quattoni2009recognizing} for scene recognition. The results are 
reported in Table \ref{tab:cnnfeatures}, where it can be seen that the 
compressed features have almost the same transfer learning performance of the 
non-compressed ones, with a drop in mean Average Precision (mAP) on Pascal VOC 2007 of $\approx\!1.0\%$ 
and a drop in accuracy on MIT-67 of $\approx\!0.2\%$.
 
\subsubsection{Fine Grained Recognition}\label{sec:fine_grained}
The second problem faced was fine grained recognition that involves recognizing 
sub-classes of the same object class such as different bird species, dog breeds, 
flower types, etc. The task is different from the one faced in Sec. 
\ref{sec:image_classification} since the differences across different 
subordinate classes are very subtle and they require a fine-detailed representation. 
We evaluated the compressed \muacro{CNN} features on two fine-grained recognition 
datasets: Caltech-UCSD Birds (CUB) 200-2011 \cite{wah2011caltech} and Oxford 102 
flowers \cite{nilsback2008automated}. The results are reported in Table 
\ref{tab:cnnfeatures}, where it can be seen that the compressed features have 
slightly better transfer learning performance of the non-compressed ones, with 
an increase in accuracy of $\approx\!1.3\%$ and $\approx\!0.7\%$ on Caltech-UCSD 
Birds and Oxford 102 flowers respectively.

\subsubsection{Attribute Detection}\label{sec:attribute}
The third problem faced was attribute detection, which in the context of computer 
vision is defined as the detection of some semantic or abstract quality shared 
by different instances/categories. We used two datasets for attribute detection: 
the UIUC 64 object attributes dataset \cite{farhadi2009describing} and the H3D 
dataset \cite{bourdev2011describing} which defines 9 attributes for a subset of 
the person images from Pascal VOC 2007. The results are reported in Table 
\ref{tab:cnnfeatures}, where it can be seen that the compressed features have 
almost the same transfer learning performance of the non-compressed ones, with a 
drop in mean Area Under Curve (mAUC) on UIUC 64 of $\approx\!0.5\%$ and a drop in mAP on H3D of 
$\approx\!0.2\%$.

\subsubsection{Visual Instance Retrieval}\label{sec:retrieval}
The fourth problem faced was visual instance retrieval, which consists in 
retrieving from a given target dataset the most similar images to a given query 
image. The similarity between images was obtained as the Euclidean distance 
between the corresponding feature vectors. The ground truth was 
defined as the set of the target database images that were relevant or not 
relevant to a given query. 
 
We considered five datasets from the state-of-the-art:

\begin{enumerate}[noitemsep,topsep=0pt, label=(\roman*)]
    \item{Oxford5k buildings} \cite{philbin2007object}: this is a collection of 
        images depicting buildings from the city of Oxford with 55 query 
        and 5063 target images. This retrieval task is quite challenging because 
        the visual appearance of Oxford buildings is very similar;
    \item{Paris6k buildings} \cite{philbin2008lost}: this is a collection of 
        images depicting buildings and monuments from the city of Paris with 55 
        query and 6412 target images. This task is less challenging 
        then the previous one because the images of the dataset are more diverse 
        than those in Oxford5k;
    \item{Sculptures6k} \cite{arandjelovic2011smooth}: This collection contains 
        6340 images of sculptures by Moore and Rodin, divided in 
        train and test (with 70 query images);
    \item{Holidays dataset} \cite{jegou2008hamming}: this collection contains 
        1491 images (with 500 query images) of different scenes, items and 
        monuments. The images are quite diverse, so this dataset is less 
        challenging than the previous ones. The performance for all the above 
        datasets was assessed by calculating the mAP;
    \item{UKbench} \cite{nister2006scalable} this collection contains 2250 
        items, each from four different viewpoints with a total of 10200 images. 
        Each image of the collection is used as a query and we assessed the 
        performance using the Recall at top four (Recall@4). 
\end{enumerate}

The results for the visual instance retrieval problem are reported in Table 
\ref{tab:cnnfeatures}. It can be seen that, also in this case, the compressed 
features have almost the same transfer learning performance of the 
non-compressed ones, with a drop in mAP on Sculptures6k and Holidays of 
$\approx\!0.1\%$. On the other three datasets instead we can observe a slight 
improvement in performance with an increase of $\approx\!4.3\%$ and 
$\approx\!0.5\%$ in mAP on Oxford5k buildings and Paris6k buildings 
respectively, and of $\approx\!1.3\%$ in Recall@4 on UKbench.

\smallskip

From all the recognition experiments considered, we note that, on average, 
there is no loss in transfer learning performance of the compressed features 
compared to the non-compressed ones. This phenomenon might be explained 
by \cite{shwartz2017opening}. In their work, the authors 
Shwartz-Ziv~and~Tishby showed that the training process of deep neural networks 
is characterized by two distinct phases: the first one consists into fast drift, 
in which the training error is reduced; the second one involves stochastic 
relaxation (\ie random diffusion) constrained by the training error value. This 
second phase leads to a decrease of the mutual information between the 
probability distributions of each layer weights and inputs, \ie an implicit compression of 
the representations.

%% file: conclusion.tex
\section{Conclusion}
We presented a method to improve the pruning step of the current 
state-of-the-art methodology to compress neural networks: \emph{Deep 
Compression} \cite{Han2015}. 
% We focused on this stage of the pipeline since 
% it was shown that quantization is orthogonal to network pruning 
% \cite{han2015learning} and it is known that pruning, quantization, and Huffman 
% coding are able to compress the network without interfering each other 
% \cite{Han2015}. 
The proposed approach is general purpose, and it can be easily applied to all 
network architectures.

The novelty of our technique lies in differentiability of the pruning phase with 
respect to the thresholds, thus allowing pruning during the 
backward phase of the learning procedure and exploiting regular gradient descent
techniques for the whole pruning phase. Moreover, since the thresholds are
learnable, the backpropagation can jointly optimize on both the network weights 
and the pruning thresholds. Furthermore, since there is a threshold per layer 
(or per filter), every layer (filter) can be optimized independently of all the 
other ones. As far as we know, this is the first approach able to 
jointly prune and learn network weights.

We showed that the proposed compression pipeline improves the current state-of-the-art
regarding pruning rate (up to $19\!\times$ compression due to pruning 
for small networks on \mnist, and $12\!\times$ compression for a big network on 
\ilsvrc), with no or negligible drop in network accuracy and strongly reducing the training time. This leads to smaller 
memory capacity and bandwidth requirements for real-time image processing, 
making it easier to be deployed on mobile systems. 

Moreover, we showed in a transfer learning scenario that the compression phase 
does not alter the effectiveness and generality of the learned representations, 
obtaining in the worst case a negligible drop in performance of $\approx\!1.2\%$ 
and in the best case an improvement of $\approx\!4.3\%$. This was verified on 
the wide range of recognition tasks identified in \cite{sharif2014cnn} on a 
diverse set of datasets: object image classification (Pascal VOC 2007), scene 
recognition (MIT-67 indoor scenes), fine grained recognition (Birds CUB 200-2011 
and Oxford 102 flowers), attribute detection (UIUC 64 objects attributes and H3D 
person attributes) and image retrieval (Oxford5k buildings, Paris6k buildings, 
Sculptures6k, Holidays dataset and UKbench).
We believe that, this is the first work where the generalization 
properties of compressed networks have been analyzed.

%% file: future.tex
\smallskip
In our opinion, interesting extensions of this work are:
\begin{enumerate*}[label=(\roman*)]
    \item further testing the pruning technique proposed in this paper by 
        considering \eg other \emph{big} networks (\muacro{VGG}) or \emph{small} 
        more recent networks (\muacro{ResNet});
    \item quantizing and Huffman coding the pruned network, so to mimic the full 
        compression pipeline proposed by \emph{Deep Compression};
    \item comparing the generalization capabilities of the networks pruned with our 
        technique with ones compressed by \emph{Deep Compression};
    \item devising a differentiable quantization technique, so to achieve a 
        pruning+quantization step learnable by stochastic gradient descent methods.
\end{enumerate*}